\documentclass[10pt,twocolumn,letterpaper]{article}

\usepackage{ijcb}
\usepackage{times}
\usepackage{epsfig}
\usepackage{graphicx}
\usepackage{amsmath}
\usepackage{amssymb}
\usepackage{multirow}

\usepackage{hyperref}

\ijcbfinalcopy % *** Uncomment this line for the final submission

\def\ijcbPaperID{73} % *** Enter the IJCB Paper ID here

\ifijcbfinal\fi
\begin{document}

%%%%%%%%% TITLE
\title{Aggregate Channel Features for Multi-view Face Detection}

\author{Bin Yang\qquad Junjie Yan\qquad Zhen Lei\qquad Stan Z. Li\thanks{Corresponding author.}\\
Center for Biometrics and Security Research \& National Laboratory of Pattern Recognition\\
Institute of Automation, Chinese Academy of Sciences, China\\
{\tt\small yb.derek@gmail.com\qquad \{jjyan,zlei,szli\}@nlpr.ia.ac.cn}
}

\maketitle
\thispagestyle{empty}

%%%%%%%%% ABSTRACT
\begin{abstract}
    Face detection has drawn much attention in recent decades since the seminal work by Viola and Jones. While many subsequences have improved the work with more powerful learning algorithms, the feature representation used for face detection still can't meet the demand for effectively and efficiently handling faces with large appearance variance in the wild. To solve this bottleneck, we borrow the concept of channel features to the face detection domain, which extends the image channel to diverse types like gradient magnitude and oriented gradient histograms and therefore encodes rich information in a simple form. We adopt a novel variant called aggregate channel features, make a full exploration of feature design, and discover a multi-scale version of features with better performance. To deal with poses of faces in the wild, we propose a multi-view detection approach featuring score re-ranking and detection adjustment. Following the learning pipelines in Viola-Jones framework, the multi-view face detector using aggregate channel features shows competitive performance against state-of-the-art algorithms on AFW and FDDB testsets, while runs at 42 FPS on VGA images.
\end{abstract}

%%%%%%%%% BODY TEXT
\section{Introduction}

Human face detection have long been one of the most fundamental problems in computer vision and human-computer interaction. In the past decade, the most influential work should be the face detection framework proposed by Viola and Jones~\cite{VJ}. The Viola-Jones (abbreviated as VJ below) framework uses rectangular Haar-like features and learns the hypothesis using Adaboost algorithm. Combined with the attentional cascade structure, the VJ detector achieved real-time face detection at that time. Despite the great success of the VJ detector, the performance is still far from satisfactory due to the large appearance variance of faces in unconstrained settings.

To handle faces in the wild, many subsequences of VJ framework merged. These methods mainly get the performance gains in two aspects, more complicated features~\cite{joint,polygon,hogCas} and (or) more powerful learning algorithms~\cite{LiMulFaceDet,softCas,MIP}. As the combination of boosting and cascade has been proven to be quite effective in face detection, the bottleneck lies in the feature representation since complicated features adopted in the above literatures bring about limited performance gains at the cost of large computation cost.

\begin{figure}[t]
\begin{center}
%\fbox{\rule{0pt}{2in} \rule{0.9\linewidth}{0pt}}
   \includegraphics[width=0.9\linewidth]{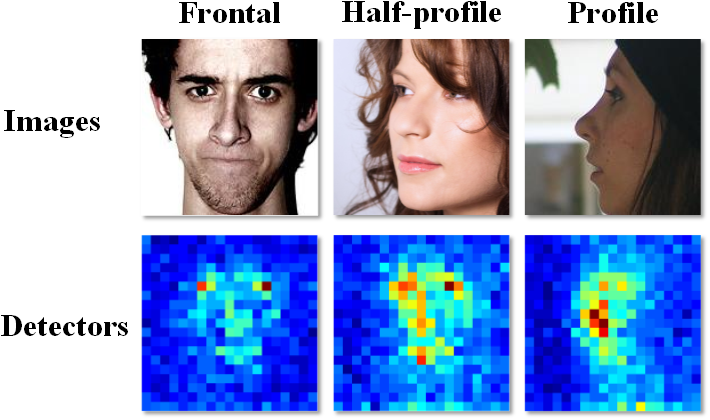}
\end{center}
   \caption{An intuitive visualization of our multi-view face detector using aggregate channel features. The area with warmer color indicates more attention paid to by the detector.}
\label{fig:highlight}
\end{figure}

Lately in another domain of pedestrian detection, a family of channel features has achieved record performances~\cite{ICF,pyramids}. Channel features compute registered maps of the original images like gradients and histograms of oriented gradients and then extract features on these extended channels. The classifier learning process follows the VJ framework pipeline. In this paper, we adopt a variant of channel features called aggregate channel features~\cite{pyramids}, which are extracted directly as pixel values on subsampled channels. Channel extension offers rich representation capacity, while simple feature form guarantees fast computation. With these two superiorities, the aggregate channel features break through the bottleneck in VJ framework and have the potential to make great advance in face detection.

As we mainly concentrate our efforts to the feature representation rather than learning algorithms in this paper, we not only just adopt the aggregate channel features in face detection, but also try to explore the full potential of this novel representation. To do so, we make a deep and all-round investigation into the specific feature parameters concerning channel types, feature pool size, subsampling method, feature scale and so on, which gives insights into the feature design and hopefully provides helpful guidelines for practitioners. Through the deep exploration, we find that: 1) multi-scaling the feature representation further enriches the representation capacity since original aggregate channel features have uniform feature scale; 2) different combinations of channel types impact the performance greatly, while for face detection the color channel in LUV space, plus gradient magnitude channel and gradient histograms channels in RGB space show best result; 3) multi-view detection is proven to be a good match with aggregate channel features as the representation naturally encodes the facial structure (Figure~\ref{fig:highlight}).

Although multi-view detection could effectively deal with diverse poses, additional issues come up as how to merge detections output by separately trained subview detectors, and how to deal with the offsets of location and scale between output detections and ground-truth. We solve these problems by carefully designed post-processing including score re-ranking, detection merging and bounding box adjustment.

The detailed experimental exploration of aggregate channel features, along with our improvements on multi-view detection, leads to large performance gain in face detection in the wild. On two challenging face databases, AFW and FDDB, the proposed multi-view face detector shows competitive performance against state-of-the-art detectors in both detection accuracy and speed.

The remaining parts of this paper are organized as follows. Section 2 revisits related work in face detection. Section 3 describes how we build the face detector using aggregate channel features. Section 4 addresses problems concerning multi-view face detection. Experimental results on AFW and FDDB are shown in section 5 and we conclude the paper in section 6.

%-------------------------------------------------------------------------
\section{Related work}
%Specifically, they use integral images to fast compute rectangular Haar-like features at various sizes and positions (constant operations per rectangular feature). The classifier is then learned with a standard Adaboost algorithm, where each weak classifier is defined as a decision stump which is constrained to rely on only one feature. The boosted classifier is structured as an attentional cascade. With this structure, an overwhelming majority of negative windows can be rejected in early stages of the detector and only the minor positives will be passed through the whole cascade. Thanks to the help of the integral image representation and the cascade structure, detection can be processed in real-time speed.

Face detection has drawn much attention since the early time of computer vision. Although many solutions had been put forward, it was not until Viola and Jones~\cite{VJ} proposed their milestone work that face detection saw surprising progress in the past decades. The VJ face detector features in three aspects: fast feature computation via integral image representation, classifier learning using Adaboost, and the attentional cascade structure. One main drawback of the VJ framework is that the features have limited representation capacity, while the feature pool size is quite large to compensate for that. Typically, in a $24\times24$ detection window, the number of Haar-like features is 160,000~\cite{VJ}. To address the problem, efforts are made in two directions. Some focus on more complicated features like HoG~\cite{hogCas}, SURF~\cite{surfCas}. Some aim to speed up the feature selection in a heuristic way~\cite{fastTrain,casDesign}. However, the problem hasn't been solved perfectly. In this paper, we mainly focus on the feature representation part and make a deep exploration into it, which is complementary to existing work on the learning algorithm and classifier structure in the VJ framework.

Recently channel features have been proposed and shown record performance in pedestrian detection~\cite{ICF,pyramids}. Due to the channel extension to diverse types like gradients and local histograms, the features show richer representation capacity for classification. However, the features are extracted as rectangular sums at various locations and scales which we believe leads to a redundant feature pool. During preparation of this paper, Mathias \etal~\cite{eccv14} independently discover the effectiveness of integral channel features in face detection domain. In this paper, we adopt a novel variant of channel features called aggregate channel features, which extract features directly as pixel values in extended channels without computing rectangular sums at various locations and scales. The feature has powerful representation capacity and the feature pool size is only several thousands. Through careful design in section 3 and implementation of multi-view detection in section 4, the aggregate channel features based detector achieves state-of-the-art performance on challenging databases.

%------------------------------------------------------------------------
\section{Proposed face detector}

In this section, we make a full exploration of the aggregate channel features in the context of face detection. We first give a brief introduction of the feature itself, including its computation, properties and advantages over traditional Haar-like features used in VJ framework. Then the detailed experimental investigation is described in two parts, feature design and training design. Before that, some guidelines concerning how we conduct the investigation are demonstrated. Each design part is divided into several separate experiments ended with a summary explaining the specific parameters used in our proposed face detector. Note that each experiment focuses on only one parameter and the others remain constant. Through the well-designed experiments, the proposed face detector based on aggregate channel features is built step by step. Issues concerning the implementation of multi-view face detection which further improves the performance are discussed in the next section.

\subsection{Feature description}

\begin{figure}[t]
\begin{center}
%\fbox{\rule{0pt}{2in} \rule{.9\linewidth}{0pt}}
  \includegraphics[width=1\linewidth]{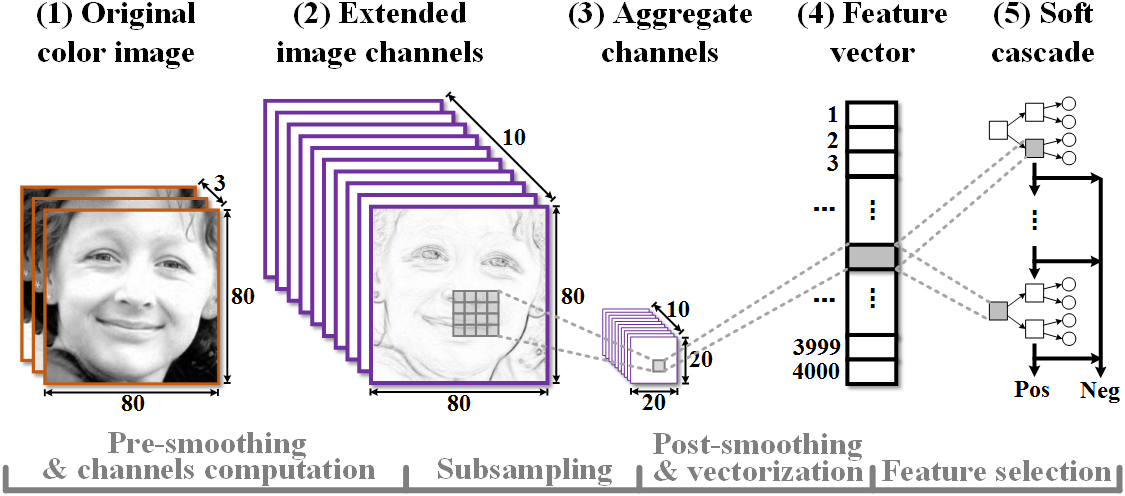}
\end{center}
   \caption{Work-flow of proposed face detector.}
\label{fig:flowchart}
\end{figure}

\textbf{Channel extension:} The basic structure of the aggregate channel features is channel. The application of channel has a long history since digital images were invented. The most common type of channel should be the color channels of the image, with Gray-scale and RGB being typical ones. Besides color channels, many different channel types have been invented to encode different types of information for more difficult problems. Generally, channels can be defined as a registered map of the original image, whose pixels are computed from corresponding patches of original pixels~\cite{ICF}. Different channels can be computed with linear or non-linear transformation of the original image. To allow for sliding window detection, the transformations are constrained to be translationally invariant.

\textbf{Feature computation:} Based on the definition of channels, the computation of aggregate channel features is quite simple. As shown in Figure~\ref{fig:flowchart}, given a color image, all defined channels are computed and subsampled by a pre-set factor. The aggregate pixels in all subsampled channels are then vectorized into a pixel look-up table. Note that an optional smoothing procedure can be done on each channel with a binomial filter both before computation and after subsampling.

\textbf{Classifier learning:} The learning process is quite simple. Two changes are made compared with VJ framework. First is that weak classifier is changed from decision stump to depth-2 decision tree. The more complex weak classifier shows stronger ability in seeking the discriminant intra and inter channel correlations for classification~\cite{akjain}. Second difference is that soft-cascade~\cite{softCas} structure is used. Unlike the attentional cascade structure in VJ framework which has several cascade stages, a single-stage classifier is trained on the whole training data and a threshold is then set after each weak classifier picked by Adaboost. These two changes lead to more efficient training and detection.

\textbf{Overall superiority:} Compared with traditional Haar-like features used in VJ framework, aggregate channel features have the following differences and advantages: 1) The image channels are extended to more types in order to encode diverse information like color, gradients, local histograms and so on, therefore possess richer representation capacity. 2) Features are extracted directly as pixel values on downsampled channels rather than computing rectangular sums with various locations and scales using integral images, leading to a faster feature computation and smaller feature pool size for boosting learning. With the help of cascade structure, detection speed is accelerated more. 3) Due to its structure consistence with the overall image, when coupled with boosting method, the boosted classifier naturally encodes structured pattern information from large training data (see Figure~\ref{fig:highlight} for an illustration), which gives more accurate localization of faces in the image.

\subsection{Investigation guidelines}

All investigations are trained on the AFLW face database\setcounter{footnote}{0}\footnote{\url{http://lrs.icg.tugraz.at/research/aflw/}}~\cite{AFLW} and tested on the Annotated Faces in the Wild (AFW) testset\footnote{\url{http://www.ics.uci.edu/~xzhu/face/}}. To make it clear, there are in total $36,112$ positive samples and $108,336$ negative samples selected from AFLW which are kept constant in all investigations. Testset contains $205$ natural images with faces that vary a lot in pose, appearance and illumination.

To alleviate the ground-truth offset caused by different annotation styles (Figure~\ref{fig:view}) in training and testing set and make the evaluation more comparable, a lower Jaccard index\footnote{The Jaccard index is defined as the size of the intersection divided by the size of the union of the sample sets.} with threshold $0.3$ is adopted in comparative evaluation. Practically the lower threshold won't cause errors being mistakenly corrected. Note that in final evaluation of the proposed face detector (section 5), the AFW testset, together with another face benchmark FDDB database, are used as testbed and the evaluation metric follows the database protocol.

\subsection{Feature design}

\begin{figure*}
\begin{center}
%\fbox{\rule{0pt}{2in} \rule{.9\linewidth}{0pt}}
  \includegraphics[width=0.95\linewidth]{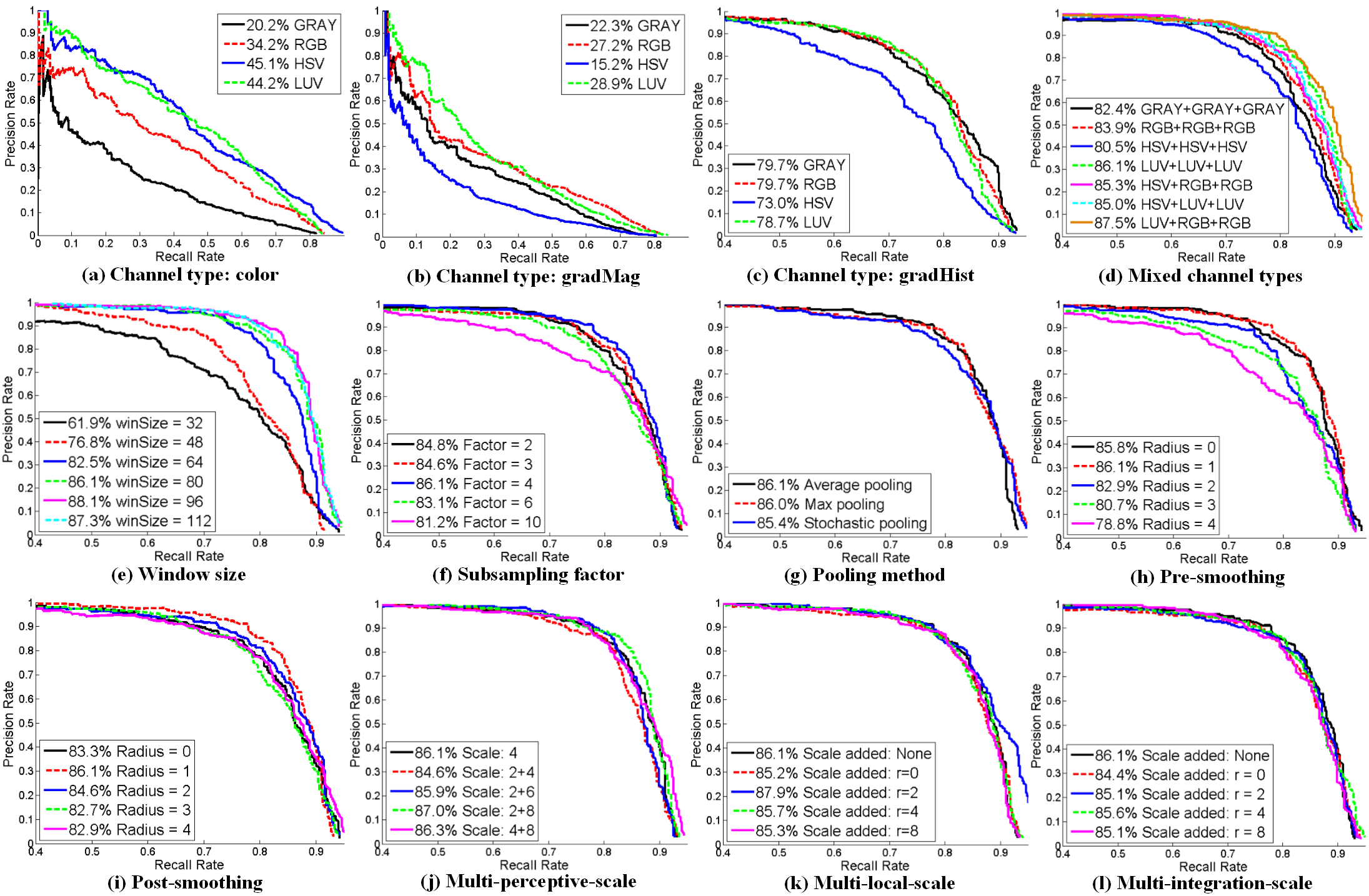}
\end{center}
   \caption{Results of comparative experiments in feature design. Best viewed in color.}
\label{fig:channel}
\end{figure*}

To fully exploit the power of aggregate channel features in face detection domain, a deep investigation into the design of the feature is done mainly on channel types, window size, subsampling method and feature scale. Results of comparative experiments are shown in Figure~\ref{fig:channel}.

\textbf{Channel types:} Three types of channels are used, which are color channel (Gray-scale, RGB, HSV and LUV), gradient magnitude, and gradient histograms. The computation of the latter two channel types could be seen as a generalized version of HoG features. Specifically, gradient magnitude is the biggest response on all three color channels, and oriented gradient histograms follow the idea of HoG in that: 1) rectangular cell size in HoG equals the subsampling factor in aggregated channel features; 2) each orientation bin results in one feature channel (6 orientation bins are used in this paper). Figure~\ref{fig:channel} (a)\~{}(c) show how much each of these three types alone contributes to the performance of face detection. It can be seen that the gradient histograms contribute most to the performance among all three channel types. Figure~\ref{fig:channel} (d) shows the performances of combinations of these three types computed on different color channels.

\textbf{Detection window size:} Detection window size is the scale to which we resize all face and non-face samples and then train our detector. Larger window size includes more pixels in feature pool and thus may improve the face detection performance. On the other hand, too large window will miss some small faces and diminish the detection efficiency. Figure~\ref{fig:channel} (e) shows comparison of window size ranging from $32$ to $112$ with a stride of $16$ pixels.

\textbf{Subsampling:} The factor for subsampling can be regarded as the perceptive scale for that it controls the scale at which the aggregation is done. Changing the factor from large to small leads to the feature representation shifting from coarse to fine and the feature pool size getting bigger. Experiments on different subsampling factors are shown in Figure~\ref{fig:channel} (f). In original aggregate channel features, the way to do subsampling is average pooling. Following the idea in Convolutional Neural Networks, another two ways of subsampling, max pooling and stochastic pooling~\cite{stoPooling} are tested in Figure~\ref{fig:channel} (g).

\textbf{Smoothing:} As described in feature description, both pre and post smoothing is done in default setting of aggregate channel features. A binomial filter with a radius of $1$ is used for smoothing. The smoothing procedure also has a great influence on the scale of the feature representation. Concretely, pre-smoothing determines how far the local neighborhood is in which local correlations are encoded before channel computation, while post-smoothing determines the neighborhood size in which the computed channel features are integrated with each other. In~\cite{ICF}, the former corresponds to the `local scale' of the feature, while the latter represents the `integration scale'. We vary the filter radius used in pre and post smoothing and find that both using a radius of $1$ gets the best results. Figure~\ref{fig:channel} (h)\~{}(i) present the comparative results.

\textbf{Multi-scale:} In aggregate channel features, although hidden information at different scale could be extracted at a cost of more weak classifiers, it would be better to make the integrated channel features multi-scaled and thus make themselves more discriminant. Therefore the same or better classification performance can be achieved with fewer weak classifiers. In this part, we implement three multi-scale version of aggregate channel features in the aforementioned three kinds of scale, perceptive scale (subsampling), local scale (pre-smoothing) and integration scale (post-smoothing) and compare their performaces. See results in Figure~\ref{fig:channel} (j)\~{}(l).

\textbf{Summary:} The color channel, gradient magnitude and gradient histograms prove themselves a good match in aggregate channel features. However, different choices of color channel used and on which gradients are computed have a great impact on performance. According to the experiments, LUV channel and gradient magnitude and 6-bin histograms computed on RGB color space (in total 10 channels) are the best choice for face detection.

Larger detection window size generally gets better performance, but will miss many small faces in testing and lead to inefficient detection. In this work, we set the size to $80\times80$ as its optimal performance.

A subsampling factor of $4$ is most reasonable according to the experiments, while different pooling methods show small differences. However, max pooling and stochastic pooling are much slower than average pooling, therefore the average pooling becomes the best match for the sake of efficiency. In this way, the resulting feature pool size of our face detector is $(80/4)\times(80/4)\times10=4000$, considerably smaller than that in VJ framework.

As for multi-scale version of aggregate channel features, multi-local-scale with an additional scale of radius $2$ shows the best performance. The probable reason is that pre-smoothing controls the local scale of the neighborhood feature correlations and therefore matches the intuition inside multi-scale best. Compared with other fine-tuning, the multi-scale version has a notable performance gain for that it makes up for the scale uniformity caused by subsampling to some extent. One main drawback is that it doubles the feature pool size and as a result slows down the detection speed somewhat. Based on the trade-off, we implement two face detectors with different scale settings, one is single-scaled with faster speed and the other is multi-scaled with better accuracy. We evaluate and discuss the performances of these two versions in detail in section 5.

%------------------------------------------------------------------------
\subsection{Training design}

Besides careful design of the aggregate channel features, experiments on the training process which is similar to that in VJ framework are also carried out. The differences are that the weak classifier is changed into depth-2 decision tree and soft-cascade~\cite{softCas} structure is used. Details of the training design are as follows.

\begin{figure}[t]
\begin{center}
%\fbox{\rule{0pt}{2in} \rule{0.9\linewidth}{0pt}}
   \includegraphics[width=0.9\linewidth]{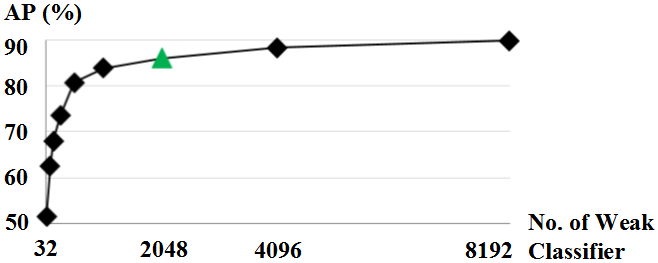}
\end{center}
   \caption{Comparison of different numbers of weak classifier in the soft cascade.}
\label{fig:clfLearn}
\end{figure}

\textbf{Number of weak classifiers:} Given a feature pool size of $4,000$, we vary the number of weak classifiers contained in the soft-cascade. In Figure~\ref{fig:clfLearn} performances of various numbers of weak classifiers ranging from $32$ to $8192$ are displayed, which shows that apparently more classifiers generate better performance, and when the number gets larger the performance begins to saturate. Since more classifiers slow down the detection speed, there's a trade-off between accuracy and speed. Searching for the saturate point as the optimal is significant during training in such framework.

\textbf{Training data:} Empirically, more training data will get better performance given powerful representation capacity. In this case, AFLW database is used as the only positive training data. However, as images in AFLW database are very salient and the background has very less variance, negative samples cropped from the AFLW database can't represent the real world scenario well, which limits the face detection performance in the wild. In this part, we further use PASCAL VOC $2007$ database and randomly crop windows from images without person as the new negative samples. Experiments show that the new training data containing cluttered background significantly improve the performance with $4.1\%$.

\textbf{Summary:} Based on observations above, we choose $2048$ as the number of weak classifiers contained in the soft cascade. As each weak classifier is a depth-2 decision tree, it takes only two comparing operations to apply a weak classifier, which is quite fast. During training, as negative data is large, we adopt a standard Bootstrap procedure to sample hard negative samples from PASCAL VOC $2007$ in the implementation of the proposed face detector.

%------------------------------------------------------------------------
\section{Multi-view detection}

Human faces in real world usually have highly varied poses. In AFLW database, the human pose is divided into three aspects: 1 in-plane rotation `roll' and 2 out-of-plane rotations `yaw' and `pitch'. Because of this large variance in face pose, it is difficult to train a single view face detector to handle all the poses effectively. A multi-view detection is further examined in this part. Due to the adoption of soft-cascade structure, a multi-view version of face detector won't cause too much computation burden. Typically, we divide the out-of-plane rotation ¡°yaw¡± into different views and let the classifier itself tolerate the pose variance in the other two types of rotations.

\begin{figure}[t]
\begin{center}
%\fbox{\rule{0pt}{2in} \rule{0.9\linewidth}{0pt}}
   \includegraphics[width=1\linewidth]{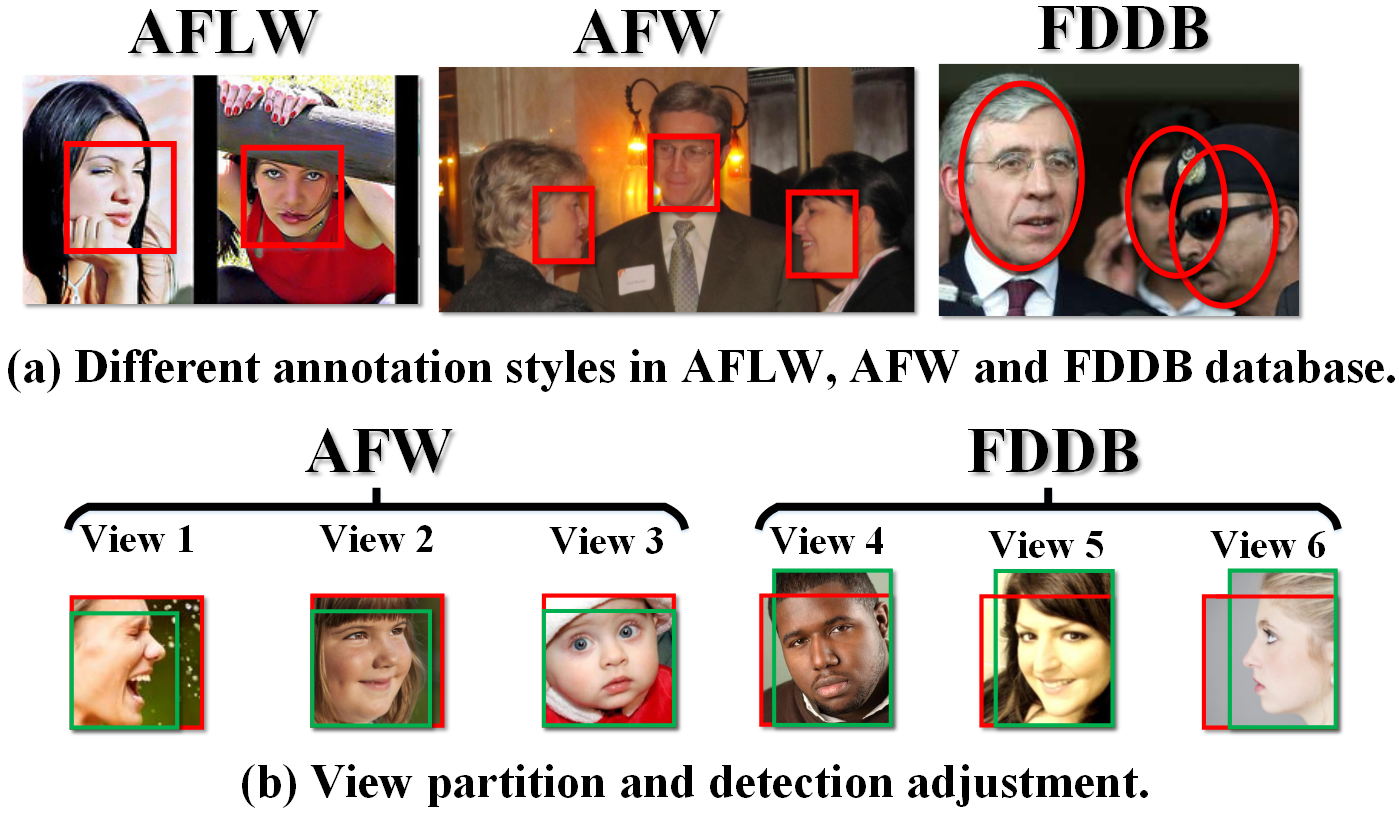}
\end{center}
   \caption{Illustration of different ground-truth annotation styles in databases, the view partition and symmetric detection adjustment. Rectangles with red and green color correspond to detections before and after adjustment.}
\label{fig:view}
\end{figure}

Adopting multi-view detection also brings about many troublesome issues. If handled improperly, the performance will differ greatly. First, detectors of different view will each produce a set of candidate positive windows followed with a set of confidence scores. For application purpose, we need to merge these detections from different views and also remove duplicated windows. A typical approach is Non-Maximum Suppression (NMS)~\cite{hog}. An issue rises on how to compare confidence scores from different classifiers and how to do window merging in the trade-off between high precision rate and high detection rate. Second, as for detection evaluation, usually the overlap of bounding boxes is used as the criterion. However, annotations in different data sets may not have a consistent style (Figure~\ref{fig:view} (a)). This diversity suffers more in profile faces. Since our face detector is trained and tested on different data sets, this issue impacts the performance a lot. Third, detectors of different views need to be trained with different samples separately. How to divide the views therefore becomes another concerning problem. In this section, we address the above three issues successfully by careful designs and therefore fully exploit the advantage of multi-view detection.

%------------------------------------------------------------------------
\subsection{View partition}

In the scenario of detecting faces in the wild, pose variation caused by yaw is usually severer than pitch and roll. Therefore we divide the faces in AFLW database according to yaw angle. We have $6$ subviews which are horizontally symmetric (see Figure.~\ref{fig:view} (b)) because we flip each image in the training set. Specifically, there are $6630$, $8446$, $9610$, $9610$, $8446$, $6630$ images in views from $1$ to $6$. Benefitting from the symmetry of our model, we can only train three subview detectors of the right side for simplicity, and use these trained right-side detectors to generate the left-side detectors. Detections of all six detectors are then merged to get the final detections. Though multi-view detection significantly improves the detection performance (especially the recall rate), the post-processing of detections from different detectors becomes a trouble. If handled improperly, the performance degrades a lot.

%------------------------------------------------------------------------
\subsection{Post-processing}

Difficulties in the post-processing of multi-view detection mainly reflect on the following aspects: 1) different score distributions and; 2) different bounding box styles. Concretely, as each subview detector is trained separately, their output confidence scores usually have different distributions. What's more, due to the annotation rule in the AFLW database that the face's nose is approximately at the center location of the bounding box ground-truth, as the subview changes, the bounding box shifts. This bounding box offset causes difficulty both in detection merging and final evaluation using Jaccard index metric. To solve these annoying issues and make the best use of multi-view detection, we introduce the following methods for post-processing.

\textbf{Score re-ranking:} We propose the following three kinds of score re-ranking: 1) normalizing scores of different views to [0, 1]; 2) defining a new score that has uniform distribution and; 3) taking overlapping detections into consideration.

\noindent$Normalization$: After training a classifier, calculate the output range of the classifier and use the range to do normalization later so that output score has a range of [0, 1].

\noindent$NewScore$: Originally, each weak classifier in the soft-cascade owns a score and final score is the sum of all scores. Instead, we use the number of weak classifier that the image patch passed positively as the new score. Therefore the upper limit of the new score is $2048$ in our case.

\noindent$OverlapRerank$: Given an image, multiple detections from multi-view detectors exist each with a score. For each detection, we first calculate the number of overlapped detection it has (overlap threshold is $0.65$) and then multiply score of each detection with a factor of its overlapping number ranking\setcounter{footnote}{0}\footnote{A toy example: Det1: score: $10$, nOverlap: $10$; Det2: score: $9$, nOverlap: $20$; Det3: score: $5$, nOverlap: $5$. After score re-ranking: Det1: score: $10\times\frac{2}{3}=6.67$; Det2: score: $9\times\frac{3}{3}=9$; Det3: score: $5\times\frac{1}{3}=1.67$.}.

\noindent$SumofOverlap$: Instead of using overlapping as a multiply factor, here we use the sum of overlapped detections' scores as the current detection's new score.

\textbf{Detection merging:} Apart from the $Greedy*$ version of Non Maximum Suppression~\cite{ICF}, we also use the detection combination introduced in~\cite{overfeat}. It averages the locations of overlapped detections rather than suppresses them.

\textbf{Detection adjustment:} As shown in Figure~\ref{fig:view} (a), different databases have different annotation styles of ground-truth. Specifically, AFLW has square annotations with nose located approximately at the center. AFW uses tight rectangular bounding boxes as annotations with the eye-brow being the approximate upper bound. FDDB uses elliptical annotations bounding the whole head. As our detector is trained on AFLW and tested on AFW and FDDB, there exist offsets in both detection position and scale. According to observations, the offsets vary as face pose changes. Therefore we adopt a view-specific detection adjustment to alleviate the offsets. Note that the adjustment is constant for all images and faces in the same database, see Figure~\ref{fig:view} (b) for details.

\textbf{Summary:} According to experimental results (Table~\ref{tab:reranking}), $OverlapRerank$ seems to be the best score re-ranking method. The underlying reason may be that true positives usually have many overlapped detections, while the false positives would only get a few responses. Therefore leveraging this overlapping information in score re-ranking can reduce many false positives. However, in practice, overlap related methods and detection combination both cost much time to process, which is infeasible in a large majority of applications. We finally adopt $Normalization$ score re-ranking combined with $Greedy*$ Non Maximum Suppression for the sake of detection speed.

\begin{table}
\begin{center}
\begin{tabular}{|l|c||l|c|}
\cline{1-2}\cline{3-4}
Reranking & AP (\%) & Merging & AP (\%)\\
\hline\hline
None & 91.7 & Greedy* NMS & 91.7\\
\hline
Normalization & 93.5 & \multirow{4}{*}{Combination} & \multirow{4}{*}{93.4}\\
NewScore & 92.9 & &\\
OverlapRerank & 95.0 & &\\
SumofOverlap & 93.7 & &\\
\hline
\end{tabular}
\end{center}
\caption{Comparison of different methods of score re-ranking and detection merging.}
\label{tab:reranking}
\end{table}

\begin{figure*}
\begin{center}
%\fbox{\rule{0pt}{2in} \rule{0.9\linewidth}{0pt}}
   \includegraphics[width=1\linewidth]{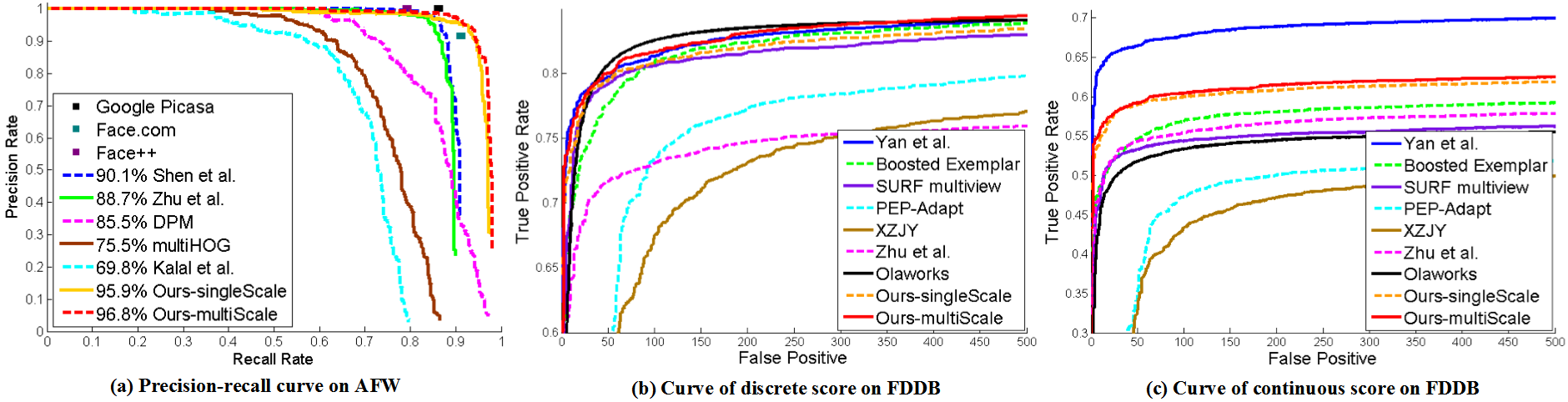}
\end{center}
   \caption{Experimental results on AFW and FDDB database. Best viewed in color.}
\label{fig:result}
\end{figure*}

%------------------------------------------------------------------------
\section{Experiments}

In this section, we compare our method with state-of-the-art methods on AFW and FDDB databases which contain challenging faces in the wild. In AFW, we compare with three commercial systems (\verb=Google Picasa=, \verb=Face.com= and \verb=Face++=) and five academic methods (Shen \etal~\cite{shen}, Zhu \etal~\cite{AFW}, DPM~\cite{DPM}, multiHOG~\cite{AFW} and Kalal \etal~\cite{kalal}). In FDDB, we compare with one commercial system (\verb=Olaworks=) and six academic methods (Yan \etal~\cite{jjyan}, Boosted Exemplar \etal~\cite{exemplar}, SURF multiview~\cite{surfCas}, PEP-Adapt~\cite{pep}, XZJY~\cite{shen} and Zhu \etal~\cite{AFW}) listed on FDDB results page\setcounter{footnote}{0}\footnote{\url{http://vis-www.cs.umass.edu/fddb/results.html}}.

%-------------------------------------------------------------------------
\subsection{Evaluation on benchmark face database}

As shown in Figure~\ref{fig:result}, in AFW, our multi-scale detector achieves an ap value of $96.8\%$, outperforming other academic methods by a large margin. When it comes to commercial systems, ours is better than \verb=Face.com= and almost equal to \verb=Face++= and \verb=Google Picasa=. Note that most of our false positives on AFW database are faces that haven't been annotated (small, seriously occluded or artificial faces like mask and cartoon character).

When evaluated on FDDB database, we follow the evaluation protocol in~\cite{FDDB} and report the average discrete and continuous ROC of the ten subfolders. For equality, we fix the number of false positives to $284$ (equivalent to an average of $1$ False Positive Per Image) and compare the true positive rate. In discrete score where evaluation metric is the same as in AFW, our detector achieves $83.7\%$, which is a little better than Yan \etal~\cite{jjyan}. Note that the ground-truth in FDDB are elliptical faces, therefore the evaluation metric of an overlap ratio bigger than $0.5$ cannot reveal the true performance of the proposed detector well. When using continuous score which takes the overlap ratio as the score, our method gets $61.9\%$ true positive rate at $1$ FPPI for multi-scale version, surpassing other methods which output rectangular detections by a notable margin (the Yan \etal detector outputs the same elliptical detections as the ground-truth, therefore having advantages with this metric). Our detector using single-scale features performs a little worse with the benefit of faster detection speed.

%-------------------------------------------------------------------------
\subsection{Discussion}

\textbf{Training efficiency:} We implement the method with Piotr's MATLAB toolbox~\cite{toolbox} on a PC with Intel Core i7-3770 CPU and 16GB RAM. With $21,328$ positive images and $5,771$ negative images in total 6 views, the training process takes about $5.3$ mins for a single-scale subview detector containing $2048$ weak classifiers and $10.2$ mins for multi-scale version. Note that we use much fewer training data than SURF multiview~\cite{surfCas} whilst still outperforming their performance.

\textbf{Comparative results:} When inspecting detections of the proposed face detector and other algorithms on the testsets, some patterns can be found to explain why our detector outperforms others. One evident strength lies in detecting faces with extreme poses. Because we adopt multi-view detection and train each subview detector separately, our detector handles pose variations very well. Second is the outstanding illumination invariance of our detector, which is mainly owing to the extension of channel types to LUV color space and gradient-related channels.
%With a soft-cascade containing $2048$ weak classifiers, our detector outperforms state-of-the-art academic methods on two challenging databases by a considerable margin, while is competitive to the commercial systems, which proves the great representation capacity of the aggregate channel features in face detection domain.

\textbf{Detection speed:} Due to the simple form of aggregate channel features and fast computation of feature pyramid~\cite{pyramids}, detection is quite efficient. For full yaw pose face detection in VGA image, the proposed detector using single-scale features runs at $20$ FPS on a single thread and $62$ FPS if $4$ threads are used. If only frontal faces are concerned, the detector runs at $34$ FPS and $95$ FPS after parallelization. When it comes to the proposed detector using multi-scale features, the above four indices reduce to $15$, $42$, $21$ and $55$ FPS. Considering the large performance gain and similar speed, the proposed method can replace Viola-Jones detector for face detection in the wild.

%------------------------------------------------------------------------
\section{Conclusion}

A novel feature representation called aggregate channel features possesses the merits of fast feature extraction and powerful representation capacity. In this paper, we successfully apply the feature representation to face detection domain through a deep investigation into the feature design, and propose a multi-scale version of feature which further enriches the representation capacity. Combined with our efforts into solving issues concerning multi-view detection, the proposed multi-view face detector shows state-of-the-art performance in both effectiveness and efficiency on faces in the wild. The proposed method appeals to real world application demands and has the potential to be embedded into low power devices.

%------------------------------------------------------------------------
\section*{Acknowledgement}

This work was supported by the Chinese National Natural Science Foundation Projects \#61105023, \#61103156, \#61105037, \#61203267, \#61375037, National Science and Technology Support Program Project \#2013BAK02B01, Chinese Academy of Sciences Project No. KGZD-EW-102-2, and AuthenMetric R\&D Funds.

{\footnotesize
\bibliographystyle{ieee}
\bibliography{bib}
}

\end{document}